\documentclass[runningheads]{llncs}

\usepackage{booktabs}
\usepackage[T1]{fontenc}

\usepackage{graphicx,verbatim}

\usepackage{hyperref}
\usepackage{url}
\usepackage{color}

\usepackage{multirow}
\usepackage{amssymb}

\begin{document}

\title{MMAP:  Multimodal Missing-Aware Pretraining for Longitudinal Alzheimer’s Prediction} 
\titlerunning{MMAP}

\author{
Fiona Kekwick\inst{1} \and
Matthew Baugh\inst{1} \and
Bernhard Kainz\inst{1,3} \and
Paul M. Matthews\inst{1,2,4} \and
Wenjia Bai\inst{1,2} 
\textnormal{for the Alzheimer's Disease Neuroimaging Initiative}\textsuperscript{*}
}

\authorrunning{F. Kekwick et al.}

\institute{
Department of Computing, Imperial College London, London, UK
\and
Department of Brain Sciences, Imperial College London, London, UK
\and
FAU Erlangen-Nürnberg, Erlangen, Germany
\and
Rosalind Franklin Institute, Didcot, UK
\\
\email{f.kekwick23@imperial.ac.uk}
}

\maketitle   
\begingroup
\renewcommand{\thefootnote}{*}
\footnotetext{\href{http://adni.loni.usc.edu/wp-content/uploads/how_to_apply/ADNI_Acknowledgement_List.pdf}{ADNI investigators} contributed to data collection but not this analysis.}
\endgroup

\begin{abstract}
Clinical decision making heavily relies on predicting the disease progression trajectory by seeking to understand patient's health status which is characterised by multimodal medical data. AI holds great potential for learning useful representations from multimodal medical data to predict disease progression and aid clinical decision making. However, development of predictive AI models is constrained by missing modalities and incomplete tabular data frequently occurring in medical datasets. In addition, disease labels alone may only provide limited supervisory signals for learning representations from high-dimensional multimodal data. Here, we present MMAP, a novel Multimodal Missing-aware Alignment Pretraining method for learning image-tabular representations from incomplete data. An image encoder is pretrained with efficient sigmoid contrastive learning combined with generative reconstruction. A tabular encoder is built upon a tabular foundation model. A missing token generator enables the two encoders to take incomplete data as input, enabling the model to be robust against missing modalities, either with missing images or missing tabular data. We evaluate the clinical usefulness of the learnt multimodal representations on two challenging longitudinal clinical tasks for Alzheimer’s disease: predicting disease stage conversion and predicting amyloid status. The proposed method outperforms strong multimodal and unimodal baselines.


\end{abstract}

\section{Introduction}

Individuals suffering from dementia caused by Alzheimer’s disease can be categorised into three stages: stable mild cognitive impairment (sMCI), progressive MCI (pMCI), and Alzheimer’s disease (AD) \cite{mueller2005AlzheimersDiseaseNeuroimaging}. Accurately predicting the transition of disease stages can improve treatment planning, and enable early intervention; ultimately improving quality of life for the patients, their families, and carers \cite{frizzell2022ArtificialIntelligenceBrain}. While deep learning models applied to medical images show promising prediction performance, recent works have shown that combining imaging scans with tabular data further improves prediction performance \cite{du2024TIPTabularImagePretraininga,hager2023BestBothWorlds,zhang2025CardiacMRIFoundationb}.

For Alzheimer’s disease assessment, magnetic resonance imaging (MRI) scans capture the structural changes of the brain and tabular data, such as cognitive test scores, assesses the performance, attention speed and memory. Multimodal models can take advantage of the different information captured in the modalities, improving diagnostic accuracy \cite{duenias2025HyperfusionHypernetworkApproach}. Nevertheless, multimodal medical data often suffers from data incompleteness: either with missing imaging scans, or with missing tabular entries. Even for well-curated studies such as the Alzheimer’s Disease Neuroimaging Initiative (ADNI)  \cite{mueller2005AlzheimersDiseaseNeuroimaging}. Consequently, multimodal models that can effectively learn from such incomplete datasets must be robust to missing data. Existing models such as \cite{duenias2025HyperfusionHypernetworkApproach,wolf2022DAFTUniversalModule} tackle the missing data issue by using bespoke architectures to learn a fused multimodal representation. However, they cannot be effectively scaled as their unique structure prevents the use of pretrained encoders. Training these models requires both supervision labels and paired image-tabular data.

To alleviate the need to train the encoders in a supervised way, pretrained unimodal encoders from foundation models offer a promising alternative which provide the base for building and finetuning a multimodal model \cite{gu2025LearningContrastiveMultimodal}. However, there are limited publicly available pretrained encoders for 3D brain MRI scans due to the computational challenge in 3D pretraining. In addition, pretrained encoders are typically not designed to handle missing data modalities.

To address these limitations, in this work, we present MMAP, a novel Multimodal Missing-aware Alignment Pretraining method that can learn from incomplete multimodal data and leverage pretrained encoders from foundation models. Our contributions are:
\begin{enumerate}
    \item An efficient sigmoid contrastive loss is combined with generative reconstruction loss, to improve the representation learning performance for the 3D brain MRI image encoder.
    \item A cross-modal missing token generator is developed, which enables the multimodal model to handle missing image data, missing tabular data, and incomplete tabular data.
    \item For two clinically challenging longitudinal tasks, disease conversion prediction and amyloid status prediction for AD, the multimodal model achieves promising prediction performance even with missing input data, performing better or on par with state-of-the-art unimodal and multimodal models.
    
\end{enumerate}

\section{Related Works}
\noindent\textbf{Image-tabular classification.} Various works have focused on improving AD classification and longitudinal prediction by using multimodal approaches to combine MRI image modality with tabular cognitive test scores \cite{duenias2025HyperfusionHypernetworkApproach,huang2023MultimodalContrastiveLearning,liu2026MANetMissingawareAttentiona,wolf2022DAFTUniversalModule}. In \cite{du2024TIPTabularImagePretraininga}, an image-tabular framework was introduced to perform multimodal downstream tasks, while being robust to missing tabular columns by using a masked tabular reconstruction task to predict missing columns. HyperFusion \cite{duenias2025HyperfusionHypernetworkApproach} and DAFT \cite{wolf2022DAFTUniversalModule} apply image-tabular learning to AD classification using the ADNI dataset \cite{mueller2005AlzheimersDiseaseNeuroimaging}. They both use an integrated early fusion approach, wherein the encoded tabular features are inserted into blocks of a 3D CNN image encoder. However, this approach may not be robust to missing modalities. Moreover, both methods use an intermediate feature-level fusion approach, which does not allow for utilising pretrained unimodal encoders. 
\\

\noindent\textbf{Contrastive multimodal pretraining.} One of the first contrastive image-tabular works was proposed by Hager et al. \cite{hager2023BestBothWorlds}, which found that aligning imaging features with tabular features improved image-only downstream task performance. In \cite{gu2025LearningContrastiveMultimodal}, a fusion module is used to fuse image and tabular features from frozen pretrained foundation models. It also introduces missing modality tokens, which are learnable tokens for missing image or tabular modalities. Contrastive learning is utilised to align multimodal embeddings of complete data with missing image features or missing tabular features~\cite{gu2025LearningContrastiveMultimodal}. This improved the performance of the model in unimodal settings and created a model robust to missing data and able to leverage existing pretrained models. 
\\

\noindent\textbf{Combining contrastive and generative pretraining.} This approach of combining contrastive and generative self-supervised pretraining has been explored in iBOT~\cite{zhou2021ibot}, which combined the contrastive alignment of the student-teacher model introduced by DINO with the generative task of masked image modelling. The iBOT loss was later incorporated into DINOv3 to improve local feature learning \cite{simeoni2025dinov3}. SigLIP2 \cite{tschannen2025SigLIP2Multilingual} adds additional generative tasks to multimodal pretraining, by adding an auto-regressive decoder to perform image-to-text captioning and referring expression prediction~\cite{tschannen2023ImageCaptionersAre,wan2024LocCaVisualPretraining}. These works demonstrate that generative tasks improve the learning for dense image features and performance in object localisation tasks.

\section{Method}
\begin{figure}[ht!]
    \centering
    \includegraphics[width=1\linewidth]{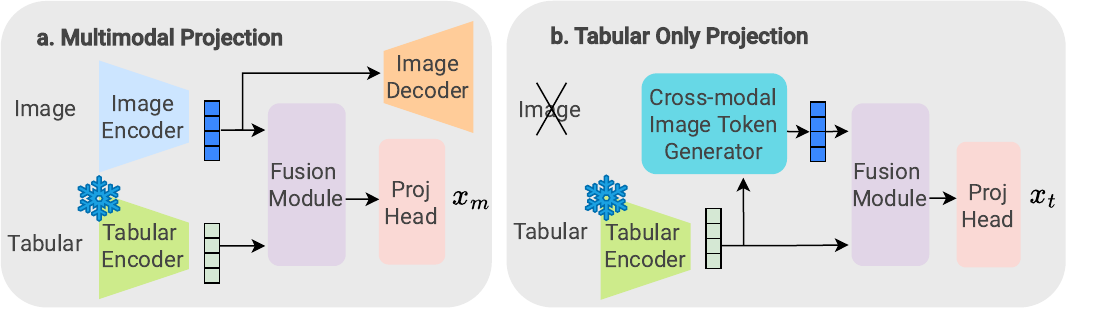}
    \caption{Architecture of MMAP, which consists of modality-specific encoders, an image decoder, a multimodal fusion module and a projection head. \textit{\textbf{a.}} Scenario when both image and tabular data are available. Image and tabular features are concatenated, fused and projected to $x_m$. \textit{\textbf{b.}} Scenario when one modality (e.g. image) is missing. A cross-modal token generator generates the image feature from available tabular features, fused and projected to tabular-only projection $x_t$.}
    \label{fig:prertaining_diagram}
\end{figure}

We propose a multimodal pretraining method, MMAP, for a model that consists of an image encoder for 3D medical images and a tabular encoder for non-imaging clinical data. Pretraining learns features from both modalities and aligns their representations. A fusion module integrates the image features with the tabular features and a projection head projects them to multimodal features $x_m$, as shown in Figure~\ref{fig:prertaining_diagram}a. In the missing modality scenario, e.g. with missing image, a tabular-to-image cross-modal token generator is used to generate the missing image features from the available tabular features, which are then fused and projected to multimodal features $x_t$, as shown in Figure~\ref{fig:prertaining_diagram}b. Similarly, if tabular data is missing, an image-to-tabular cross-modal token generator is used to generate the missing tabular features from the available image features, which are then fused and projected to multimodal features $x_i$. During pre-training, we will combine the contrastive learning loss for $x_m$, $x_i$, $x_t$ and a generative loss to reconstruct the input image via an image decoder. 
\\

\noindent\textbf{Modality encoders, image decoder and multimodal fusion.} For the image encoder, a 3D ResNet-18 architecture is used with instance normalisation \cite{he2015DeepResidualLearning}. The encoded image features are provided to an image decoder (Figure~\ref{fig:prertaining_diagram}a), comprised of five blocks of 3D convolutional LeakyReLU and upsampling layers, to perform image reconstruction. For the tabular encoder, a pretrained encoder from the tabular foundation model TabPFNv2~\cite{hollmann2025AccuratePredictionsSmall} is used. Apart from the advantage of being pre-trained on large-scale tabular data, TabPFNv2 has an inherent tabular generative model, which can predict missing entries. This allows our method to be robust to missing tabular entries, in addition to missing modalities. The fusion module, adapted from~\cite{gu2025LearningContrastiveMultimodal}, performs layer normalisation for image and tabular features, concatenates them, and passes them through an MLP. After the fusion module, the projection head projects the fused features into a joint projection space; we use a MLP projection head with 2 layers.
\\

\noindent\textbf{Combining contrastive and generative learning.} 
We employ SigLIP loss for cross-modal contrastive learning \cite{zhai2023SigmoidLossLanguage}. SigLIP loss maximises the cosine similarity between the features of two modalities. Instead of using unimodal features, we align the multimodal projected features $x_m$ with the tabular-only projection $x_t$, and with the image-only projection $x_i$. The SigLIP contrastive loss for the multimodal and tabular only projections is formulated as,
\begin{equation}
\mathcal {L}^{con}_{t,m}=\sum _{t\in N}\sum _{m\in N}\textrm{log}\frac{1}{1+e^{z_{t,m}(-\tau \mathbf{x_t} \cdot \mathbf{y_m} + b)}}.
\label{eq:contrast}
\end{equation}
where $N$ denotes the batch size, $t$ and $m$ denote the sample indices in a batch, $\mathbf{x_t}$ denotes the tabular-only projection for sample $t$, $\mathbf{y_m}$ denotes the multimodal projection for sample $m$. $z_{t,m} = 1$ if $t = m$ for paired data, otherwise $z_{t,m} = -1$. $\tau$ and $b$ are learnable hyperparameters. Unlike CLIP’s softmax contrastive loss \cite{radford2021learning}, SigLIP uses independent sigmoid losses over pairs. SigLIP loss was selected because preliminary experiments showed that SigLIP gave an improved performance over CLIP. In addition to formulating the constrastive loss $\mathcal{L}^{con}_{t,m}$, we formulate a similar  $\mathcal{L}^{con}_{i,m}$ to align multimodal projection $x_m$ with image-only projection $x_i$.

Contrastive learning focuses on learning global image-wise features. To improve the performance of local feature learning, we add a generative reconstruction loss $\mathcal{L}_{recon}$, using the image decoder to perform reconstruction. $\mathcal{L}_{recon}$ is defined as the the mean-squared error loss between reconstructed image and the input MRI image. The total pretraining loss is,
\begin{equation}
    \mathcal{L}_{pretrain} =  \lambda_{recons} \mathcal{L}_{recon} + \lambda_{con}( \mathcal{L}^{con}_{i,m} + \mathcal{L}^{con}_{t,m}), 
\end{equation}
where $\mathcal{L}^{con}_{i,m}$ and $\mathcal{L}^{con}_{t,m}$ are the contrastive losses between the image-only (missing tabular) and the multimodal projections, and the tabular-only (missing image) and multimodal projections respectively. $\lambda_{con}$ and $\lambda_{recons}$ are scalar constants.
\\

\noindent\textbf{Missing token generation.} To achieve robustness against missing modalities, previous works~\cite{LeeYLMultimodal,chen2024UnifiedModelLongitudinal,gu2025LearningContrastiveMultimodal} either utilise zero tokens to represent missing data embeddings, or utilise a learnable missing token. Here we propose a missing token generative method, generating tokens of the missing modality from the available modality, implemented as a cross-modal MLP projection head. For example, if the image is missing, its embedding will be predicted by the cross-modal token generator from tabular data, as shown in Fig.~\ref{fig:prertaining_diagram}b. Similarly, the embedding for missing tabular data will be predicted from the image. The parameters of the token generator are trained both during pretraining and finetuning.
\\

\noindent\textbf{Finetuning.} After pretraining, we take the pretrained multimodal model to perform finetuning for downstream tasks. We remove the projection head and add a classification head made from a linear leakyReLU and linear layer, added after the fusion module. For each sample, we get three different predictions, the prediction with full multimodal input data, the image-only prediction, and the tabular-only prediction. We optimise a total finetuning loss, by adding together the cross-entropy loss of the three training scenarios with equal weighting, formulated as, 
\begin{equation}
    \mathcal{L}_{finetune} = \mathcal{L}_{i} + \mathcal{L}_{t} + \mathcal{L}_{i,t},
\end{equation}
where $\mathcal{L}_{i,t}$ denotes the cross-entropy loss of the multimodal prediction compared to ground truth label, $\mathcal{L}_{i}$ denotes the cross-entropy loss of the image-only prediction, and $\mathcal{L}_{t}$ denotes the cross-entropy loss of the tabular-only prediction. We use an ensemble model, where the final prediction is a simple average of the predictions provided by the five ensemble models, using the same ensemble methodology as Hyperfusion~\cite{duenias2025HyperfusionHypernetworkApproach}.

\section{Experiments}
\noindent\textbf{Dataset.} Data were obtained from the Alzheimer's Disease Neuroimaging Initiative (ADNI) database~\cite{mueller2005AlzheimersDiseaseNeuroimaging}. We use 3D imaging and tabular data from 1,258 subjects, consisting of T1 MRI scans with corresponding tabular cognitive and clinical test scores. The MRI scans are preprocessed, following the same pipeline as \cite{levakov2020DeepLearningModel}, followed by z-score intensity normalisation and image downsampling to the size of 128$\times$128$\times$128. It is a longitudinal dataset with data being collected for each subject at multiple timepoints. In total, 5,548 MRI scans and tabular data are available, consisting of 1,442 cognitively normal (CN), 3,370 MCI, and 736 AD. The MCI group is further divided into two subgroups: subjects who have a future AD diagnosis (pMCI) and those who remain stable MCI (sMCI). The full dataset is split into 85\%/15\% for pretraining and test, in a subject-wise manner to avoid information leakage. After pretraining, the model is finetuned on a subset of pretraining data, with equally sized pMCI and sMCI groups, with a total of 2,232 scans. The finetuned model is then evaluated on the test set, which was unseen during training. \\

\noindent\textbf{Downstream evaluation tasks.} On the test set, the model is evaluated on two clinically challenging tasks: 1) predicting whether an individual with MCI will remain stable (sMCI) or progress to AD (pMCI) at a future timepoint, with longitudinal ground truth diagnoses; 2) predicting the amyloid status of the individual at a future timepoint. Longitudinal diagnoses are provided by ADNI \cite{mueller2005AlzheimersDiseaseNeuroimaging} and the code for defining the longitudinal labels from the tabular diagnoses is from \cite{ouyang2022DisentanglingNormalAging}. The amyloid status is a binary variable, indicating the presence of amyloid plaques on a positron emission tomography (PET) scan. The amyloid status is labelled as positive if the standardised uptake value ratio (SUVR) is greater than a threshold of 0.79, and otherwise labelled as negative~\cite{degenhardt2016FlorbetapirF18PET,adni-suvr}. PET scans are expensive and not widely accessible, so predicting amyloid status from MRI scans and tabular data is clinically useful for disease diagnosis and monitoring~\cite{cheng2024PredictingBrainAmyloid,ouyang2022DisentanglingNormalAging}. Both tasks are longitudinal tasks, where the model is given clinical data at the current timepoint and predicts the class at a future timepoint, which is 1 to 3 years after the input data was acquired. \\

\noindent\textbf{Implementation.} The model was pretrained for 100 epochs with $\lambda_{con} = 1$ and $\lambda_{recons} = 4$ , with a learning rate of 0.0005, weight decay of $1e^{-5}$ and a batch size of 60. It was then finetuned for 50 epochs with a learning rate of 0.0008, the same weight decay and batch size. Preliminary experiments showed that finetuning the tabular encoder, TabPFNv2, led to a performance drop. For this reason, we kept the tabular encoder frozen in all experiments.

\section{Results} Table~\ref{tab:combined_results} compares the proposed method, MMAP, to multimodal image-tabular models when both data modalities are available, including: HyperFusion \cite{duenias2025HyperfusionHypernetworkApproach}, DAFT \cite{wolf2022DAFTUniversalModule}, TIP \cite{du2024TIPTabularImagePretraininga}, SimCLR-pretrained image encoder concatenated with TabPFNv2 tabular encoder (SimCLR + Concat) and autoencoder-pretrained image encoder concatenated with TabPFNv2 tabular encoder (AE + Concat). It shows that MMAP outperforms these competing methods in the amyloid prediction task, in terms of balanced accuracy (BACC), area-under-curve (AUC), and F1 score. For the disease stage conversion prediction task, MMAP outperforms competing methods in BACC and AUC, and only underperforms in F1 score.

Because MMAP can generate features for missing data modality, it enables prediction even when only one data modality is available. Table~\ref{tab:combined_results} also compares MMAP to AE when only image is available, and compares MMAP to TabPFNv2 when only tabular data is available. It shows that MMAP outperforms unimodal models in most of the metrics, while being robust to missing modalities. We also notice that when both data modalities are available, MMAP performs better than unimodal models. This is unsurprising, as the performance gain of aligning image with tabular features has been shown in previous works \cite{gu2025LearningContrastiveMultimodal,hager2023BestBothWorlds,robinet2024DRIMLearningDisentangled,zhang2025CardiacMRIFoundationb}.
\\

\begin{table*}[t!]
\centering
\caption{Comparison of the proposed MMAP method to other multimodal image-tabular models and unimodal models, in terms of amyloid status prediction and MCI conversion (pMCI vs sMCI) prediction performance. Modality (Mod): image (I), tabular (T). Metrics reported are Balanced Accuracy (BACC), AUC, and F1 score. Experiments were run five times with random seeds, with the mean and standard deviation reported. 
For fair comparison, the DAFT implementation was taken from the same repository as \cite{duenias2025HyperfusionHypernetworkApproach}, and the TIP implementation was adapted to use a 3D ResNet-18. Bold numbers indicate the best result within each modality setting; underlined numbers indicate the best result overall. 
}
\label{tab:combined_results}
\resizebox{\textwidth}{!}{%
\begin{tabular}{l cc ccc ccc}
\toprule
\multirow{2}{*}{\textbf{Method}}
& \multicolumn{2}{c}{\textbf{Mod}}
& \multicolumn{3}{c}{\textbf{Amyloid}}
& \multicolumn{3}{c}{\textbf{MCI to AD conversion}} \\
\cmidrule(lr){2-3} \cmidrule(lr){4-6} \cmidrule(lr){7-9}
& I & T
& BACC & AUC & F1
& BACC & AUC & F1 \\
\midrule
HyperFusion \cite{duenias2025HyperfusionHypernetworkApproach}
& $\checkmark$ & $\checkmark$
& $0.893_{\pm 0.018}$ & $0.933_{\pm 0.010}$ & $0.903_{\pm 0.017}$
& $0.772_{\pm 0.015}$ & $0.839_{\pm 0.007}$ & $\textbf{0.703}_{\pm 0.020}$ \\
DAFT \cite{wolf2022DAFTUniversalModule}
& $\checkmark$ & $\checkmark$
& $0.875_{\pm 0.013}$ & $0.877_{\pm 0.064}$ & $0.902_{\pm 0.027}$
& $0.760_{\pm 0.010}$ & $0.832_{\pm 0.008}$ & $0.699_{\pm 0.017}$ \\
TIP \cite{du2024TIPTabularImagePretraininga}
& $\checkmark$ & $\checkmark$
& $0.886_{\pm 0.012}$ & $0.936_{\pm 0.011}$ & $0.912_{\pm 0.015}$
& $0.714_{\pm 0.045}$ & $0.774_{\pm 0.031}$ & $0.560_{\pm 0.078}$ \\
AE + Concat
& $\checkmark$ & $\checkmark$
& $0.892_{\pm 0.023}$ & $0.929_{\pm 0.005}$ & $0.918_{\pm 0.014}$
& $0.664_{\pm 0.027}$ & $0.713_{\pm 0.017}$ & $0.593_{\pm 0.054}$ \\
SimCLR + Concat
& $\checkmark$ & $\checkmark$
& $0.893_{\pm 0.016}$ & $0.933_{\pm 0.008}$ & $0.915_{\pm 0.012}$
& $0.739_{\pm 0.019}$ & $0.823_{\pm 0.016}$ & $0.659_{\pm 0.013}$ \\
MMAP (ours)
& $\checkmark$ & $\checkmark$
& $\underline{\textbf{0.911}}_{\pm 0.016}$ & $\underline{\textbf{0.940}}_{\pm 0.012}$ & $\underline{\textbf{0.929}}_{\pm 0.010}$
& $\underline{\textbf{0.786}}_{\pm 0.014}$ & $\underline{\textbf{0.850}}_{\pm 0.015}$ & $0.669_{\pm 0.013}$ \\
\midrule
AE \cite{ReducingDimensionalityData}
& $\checkmark$ & $\times$ 
& $0.872_{\pm 0.017}$ & $0.921_{\pm 0.004}$ & $0.909_{\pm 0.011}$
& $0.634_{\pm 0.049}$ & $0.720_{\pm 0.017}$ & $\textbf{0.568}_{\pm 0.023}$ \\
SimCLR \cite{chen2020SimpleFrameworkContrastive}
 & $\checkmark$ & $\times$
& $0.866_{\pm 0.012}$ & $0.926_{\pm 0.010}$ & $0.907_{\pm 0.014}$
& $0.661_{\pm 0.013}$ & $0.718_{\pm 0.026}$ & $\textbf{0.568}_{\pm 0.052}$ \\
MMAP (ours)
& $\checkmark$ & $\times$ 
& $\textbf{0.903}_{\pm 0.017}$ & $\textbf{0.927}_{\pm 0.010}$ & $\textbf{0.910}_{\pm 0.015}$
& $\textbf{0.690}_{\pm 0.019}$ & $\textbf{0.733}_{\pm 0.020}$ & $0.515_{\pm 0.023}$ \\
\midrule
TabPFNv2 \cite{hollmann2025AccuratePredictionsSmall}
& $\times$ & $\checkmark$ 
& $0.897_{\pm 0.010}$ & $0.925_{\pm 0.009}$ & $0.875_{\pm 0.008}$
& $0.755_{\pm 0.010}$ & $\textbf{0.844}_{\pm 0.012}$ & $\underline{\textbf{0.746}}_{\pm 0.008}$ \\
MMAP (ours)
& $\times$ & $\checkmark$ 
& $\textbf{0.903}_{\pm 0.007}$ & $\textbf{0.927}_{\pm 0.009}$ & $\textbf{0.922}_{\pm 0.007}$
& $\textbf{0.760}_{\pm 0.009}$ & $0.820_{\pm 0.017}$ & $0.666_{\pm 0.018}$ \\

\bottomrule

\end{tabular}
}
\end{table*}

\begin{table*}[h!]
\centering
\fontsize{7.8}{9.5}\selectfont
\setlength{\tabcolsep}{2pt}
\caption{Ablation studies of different loss terms and different methods of imputing the the missing modality features on the MCI to AD conversion prediction performance. Metrics are reported across 5 random seeds. I + T is the image and tabular scenario.}
\label{tab:ablation}
\begin{tabular}{@{}lccccc cc cc cc@{}}
\hline
\multirow{2}{*}{Ablation}
& \multirow{2}{*}{$\mathcal{L}_{con}$}
& \multirow{2}{*}{$\mathcal{L}_{recon}$}
& \multirow{2}{*}{$\mathcal{L}_{i}$}
& \multirow{2}{*}{$\mathcal{L}_{t}$}
& \multirow{2}{*}{$\mathcal{L}_{i,t}$}
& \multicolumn{2}{c}{I + T}
& \multicolumn{2}{c}{Image}
& \multicolumn{2}{c}{Tabular} \\
& & & & & &
BACC & AUC & BACC & AUC & BACC & AUC \\
\hline

\multicolumn{12}{@{}l}{\textit{Loss ablation}} \\
Full model
& $\checkmark$ & $\checkmark$ & $\checkmark$ & $\checkmark$ & $\checkmark$
& \textbf{0.786} & \textbf{0.850}
& \textbf{0.690} & \textbf{0.733}
& \textbf{0.760} & \textbf{0.820} \\
w/o $\mathcal{L}_{recon}$
& $\checkmark$ & $\times$ & $\checkmark$ & $\checkmark$ & $\checkmark$
& 0.768 & 0.829
& 0.673 & 0.713
& 0.757 & 0.805 \\
w/o $\mathcal{L}_{con}$
& $\times$ & $\checkmark$ & $\checkmark$ & $\checkmark$ & $\checkmark$
& 0.747 & 0.811
& 0.660 & 0.677
& 0.747 & 0.805 \\
w/o unimodal losses
& $\checkmark$ & $\checkmark$ & $\times$ & $\times$ & $\checkmark$
& 0.778 & 0.816
& 0.677 & 0.751
& 0.742 & 0.773 \\

\hline
\multicolumn{12}{@{}l}{\textit{Missing-modality token ablation}} \\
\multicolumn{6}{@{}l}{Learnable tokens \cite{gu2025LearningContrastiveMultimodal}}
& 0.767 & 0.796
& 0.659 & 0.518
& 0.708 & 0.797 \\
\multicolumn{6}{@{}l}{Zero tokens}
& 0.767 & 0.842
& 0.690 & 0.705
& 0.722 & \textbf{0.842} \\
\multicolumn{6}{@{}l}{Generated tokens (MMAP)}
& \textbf{0.786} & \textbf{0.853}
& \textbf{0.694} & \textbf{0.730}
& \textbf{0.750} & 0.833 \\

\hline
\end{tabular}
\end{table*}

\noindent\textbf{Ablation studies.} Table~\ref{tab:ablation} details the results of the ablation studies of the effect of different loss terms on the model performance in the AD conversion prediction task. The results show that combining the reconstructive loss, $L_{recon}$ with the contrastive loss, $L_{con}$ improves downstream task performance. Adding in the  $L_{recon}$ loss gives a greater increase in accuracy compared to contrastive loss only, in the image-tabular and image only scenarios (2.3\% and 2.5\%), but only a very small improvement in the tabular-only scenario (0.3\%). Combining reconstructive and contrastive losses is complementary and improves balanced accuracy on all data scenarios. We also ablate the finetuning loss when using only the multimodal $\mathcal{L}_{i,t}$ cross-entopy loss. Compared to when we combine it with the image only, $\mathcal{L}_{i}$ and tabular only $\mathcal{L}_{t}$ cross-entropy loss. This gives a small 1\% balanced accuracy improvement on the multimodal performance. But it gives a greater improvement in the unimodal, image-only and tabular-only tasks, the model learns to perform well regardless of which modalities are available at test time. This confirms that optimising for all input data scenarios during finetuning acts as a form of regularisation, consistent with finding in a prior work~\cite{gu2025LearningContrastiveMultimodal}.

\section{Conclusions}
We present a novel multimodal pretraining method, MMAP, that can perform 3D image-tabular pretraining with missing modalities. MMAP achieves the best or comparable performance on both MCI to AD conversion prediction and the prediction of future presence of amyloid plaques compared to state-of-the-art methods. Our pretraining method, which combines multimodal contrastive learning with generative reconstruction, improves the performance on downstream longitudinal prediction tasks. In clinical practice, patients without MRI scans or incomplete or missing cognitive test scores are commonplace. MMAP addresses this with the missing token generator, which enables a multimodal model to perform robust prediction tasks against missing modalities and incomplete tabular data. It shows the potential in closing the gap between multimodal prediction models and real-world clinical applications with data incompleteness. Future directions include evaluating this method on other medical image-tabular datasets and pathologies, and extending the cross-modal token generation method to settings with more than two modalities.

\begin{credits}
\subsubsection{\ackname} 

F.K. is supported by the UKRI Centre for Doctoral Training in AI for Healthcare (EP/S023283/1). M.B. is supported by the ERC project CHARMS (No.~101246053). P.M.M. acknowledges generous personal support from the Edmond J. Safra Foundation and Lily Safra, an NIHR Senior Investigator Award, Rosalind Franklin Institute, and the UK Dementia Research Institute.  W.B. acknowledges the support of EPSRC CVD-Net Programme Grant (EP/Z531297/1) and BHF New Horizons Grant (NH/F/23/70013). The authors gratefully acknowledge the HPC resources provided by NHR@FAU under the NHR projects b143dc and b180dc. NHR funding is provided by federal and Bavarian state authorities (HTA). NHR@FAU hardware is partially funded by the German Research Foundation (DFG~440719683). This work also received support from the ERC through project MIA-NORMAL (101083647). We further acknowledge the Isambard-AI AIRR, operated by the University of Bristol and funded by DSIT via UKRI and STFC [ST/AIRR/I-A-I/1023]~\cite{mcintoshsmith2024isambardai}. 

Data collection and sharing for the Alzheimer's Disease Neuroimaging Initiative (ADNI) are funded by the National Institute on Aging (NIH Grant U19 AG024904). The grantee organisation is the Northern California Institute for Research and Education. ADNI has previously received funding from the National Institute of Biomedical Imaging and Bioengineering, the Canadian Institutes of Health Research, and private-sector contributions through the Foundation for the National Institutes of Health.

\subsubsection{\discintname}

The authors have no competing interests to declare that are relevant to the content of this article. 

\end{credits}

\bibliographystyle{splncs04}
\bibliography{references}

\end{document}